\documentclass[final]{cvpr}

\usepackage{times}
\usepackage{epsfig}
\usepackage{graphicx}
\usepackage{amsmath}
\usepackage{amssymb}

\usepackage{multirow}

\DeclareMathOperator*{\argmin}{arg\,min}

\usepackage[pagebackref=true,breaklinks=true,colorlinks,bookmarks=false]{hyperref}

\def\cvprPaperID{2082} 
\def\confYear{CVPR 2021}

\begin{document}

\title{Learning Discriminative Prototypes with Dynamic Time Warping}

\author{Xiaobin Chang$^{1,2}$, Frederick Tung$^{2}$, Greg Mori$^{1,2}$\\
Simon Fraser University$^1$, Borealis AI$^2$\\
{\tt\small xiaobin\_chang@sfu.ca frederick.tung@borealisai.com mori@cs.sfu.ca}
}


\maketitle

\begin{abstract}
   Dynamic Time Warping (DTW) is widely used for temporal data processing. However, existing methods can neither learn the discriminative prototypes of different classes nor exploit such prototypes for further analysis.
   We propose Discriminative Prototype DTW (DP-DTW), a novel method to learn class-specific discriminative prototypes for temporal recognition tasks.
   DP-DTW shows superior performance compared to conventional DTWs on time series classification benchmarks\footnote{Code available at \url{https://github.com/BorealisAI/TSC-Disc-Proto}}.
   Combined with end-to-end deep learning, DP-DTW can handle challenging weakly supervised action segmentation problems and achieves state of the art results on standard benchmarks.
   Moreover, detailed reasoning on the input video is enabled by the learned action prototypes. Specifically, an action-based video summarization can be obtained by aligning the input sequence with action prototypes.
\end{abstract}

\section{Introduction}\label{Sec:Intro}
Temporal data is a common data form and widely exists in different domains~\cite{fu2011review}, e.g., finance, industrial processes and video sequences. Analyzing temporal sequences is thus an important task. However, a significant challenge arises when comparing two sequences as they are not guaranteed to be aligned. They can be varied in temporal length and/or observation speed. Therefore, alignment is essential before comparisons, e.g., computing their discrepancy value.  Naive pre-processing such as interpolation, cyclic repeat extension, place-holder insertion and down-sampling are used to align sequences with different lengths. These methods either modify the original data distribution or suffer from data loss and fail to handle the issue of varied speeds.

Dynamic Time Warping (DTW)~\cite{sakoe1978dynamic, berndt1994using} was proposed to handle misalignment issues in temporal data. The optimal monotonic alignment between two input sequences is provided by a dynamic programming procedure. DTW is thus robust to inputs with varied temporal lengths and observation speeds. The discrepancy value between the two sequences can then be computed based on the alignment.
With DTW and its discrepancy, a prototype over a set of sequences can be obtained by averaging.  This technique is known as DTW barycenter averaging (DBA)~\cite{petitjean2014dynamic} and enables several tasks, e.g., clustering and classification.
However, DBA considers the intra-class samples only and neglects the inter-class ones in learning the class-specific prototypes for time series classification (TSC). Discriminative prototypes thus fail to be obtained and classification performance is negatively affected.

Besides the conventional multi-class single label TSC setting~\cite{dau2019ucr}, we focus on solving the weakly supervised action segmentation problem in video data~\cite{kuehne2014language, bojanowski2014weakly}. Three major challenges are highlighted.
First, the video data is captured from realistic scenarios, such as daily activities and movies. It thus contains sophisticated spatial-temporal dynamics.
Secondly, multiple actions are performed sequentially in each video.
Last but not least, instead of labelling the action at each frame, only the action order is provided as weak supervision.
To better handle the complex temporal structure of video inputs, deep models are widely adopted to extract frame-wise deep feature representations.
However, training deep models with such weak supervision is not straightforward. Existing methods~\cite{richard2018neuralnetwork, ding2018weakly, chang2019d3tw, li2019weakly} follow very similar paradigms. Specifically, deep models first provide the action predictions of each frame. Different algorithms are then proposed to encode the frame-wise predictions with the given action transcript and result in different learning objectives.
For example, the dynamic programming procedure of DTW is exploited by D$^3$TW~\cite{chang2019d3tw} as its encoding algorithm. To obtain a differentiable DTW loss for deep learning, a relaxation technique, as in Soft-DTW~\cite{cuturi2017soft} can be adopted.
No existing work attempts to learn discriminative prototype sequences of different actions nor use them for action segmentation.

\begin{figure}[t]
\begin{center}
   \includegraphics[width=0.95\linewidth]{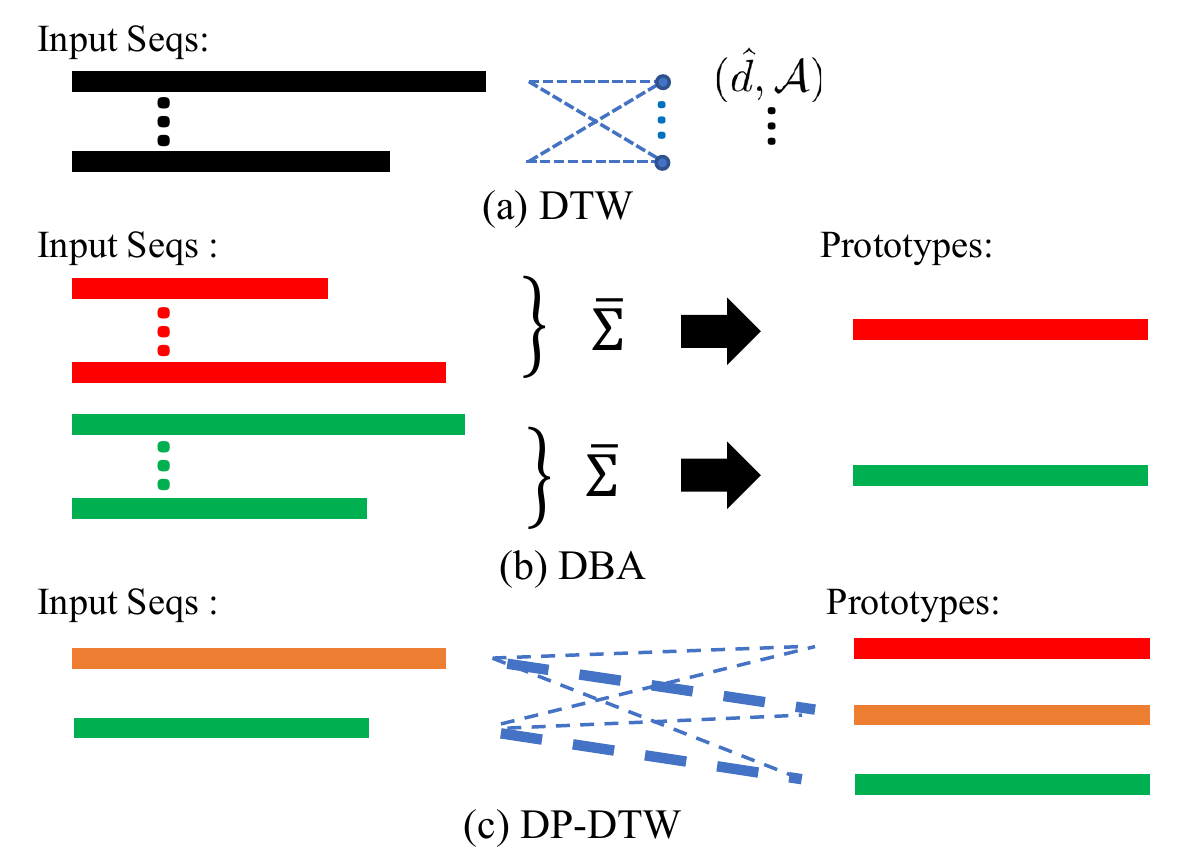}
\end{center}
   \caption{Bold lines represent sequences with varied temporal lengths. (a) DTW computes the discrepancy $\hat{d}$ and alignment $\mathcal{A}$ between a pair of sequences. (b) DBA computes a prototype by averaging (denoted as $\bar{\Sigma}$) the samples within a class. Different classes are indicated by colors. (c) DP-DTW focuses on the inter-class variance. Each input should be closest (shown as bold dashed line) to the prototype sequence of the same class (color).}
\label{fig:TSC_DTWs}
\end{figure}

In this paper, we propose a novel DTW method, Discriminative Prototype DTW (DP-DTW), for temporal recognition problems.
In the TSC setting, each sequence corresponds to a single class.
Instead of averaging the sequences within a class as in DBA~\cite{petitjean2014dynamic}, DP-DTW computes the discrepancies between an input and the prototypes of different classes and then is supervised by discriminative loss.  Class-specific distinctive temporal dynamics are thus represented by such learned prototypes.  Illustrations of DTW, DBA and the proposed DP-DTW are shown in Figure~\ref{fig:TSC_DTWs}.

\begin{figure}[t]
\begin{center}
   \includegraphics[width=0.99\linewidth]{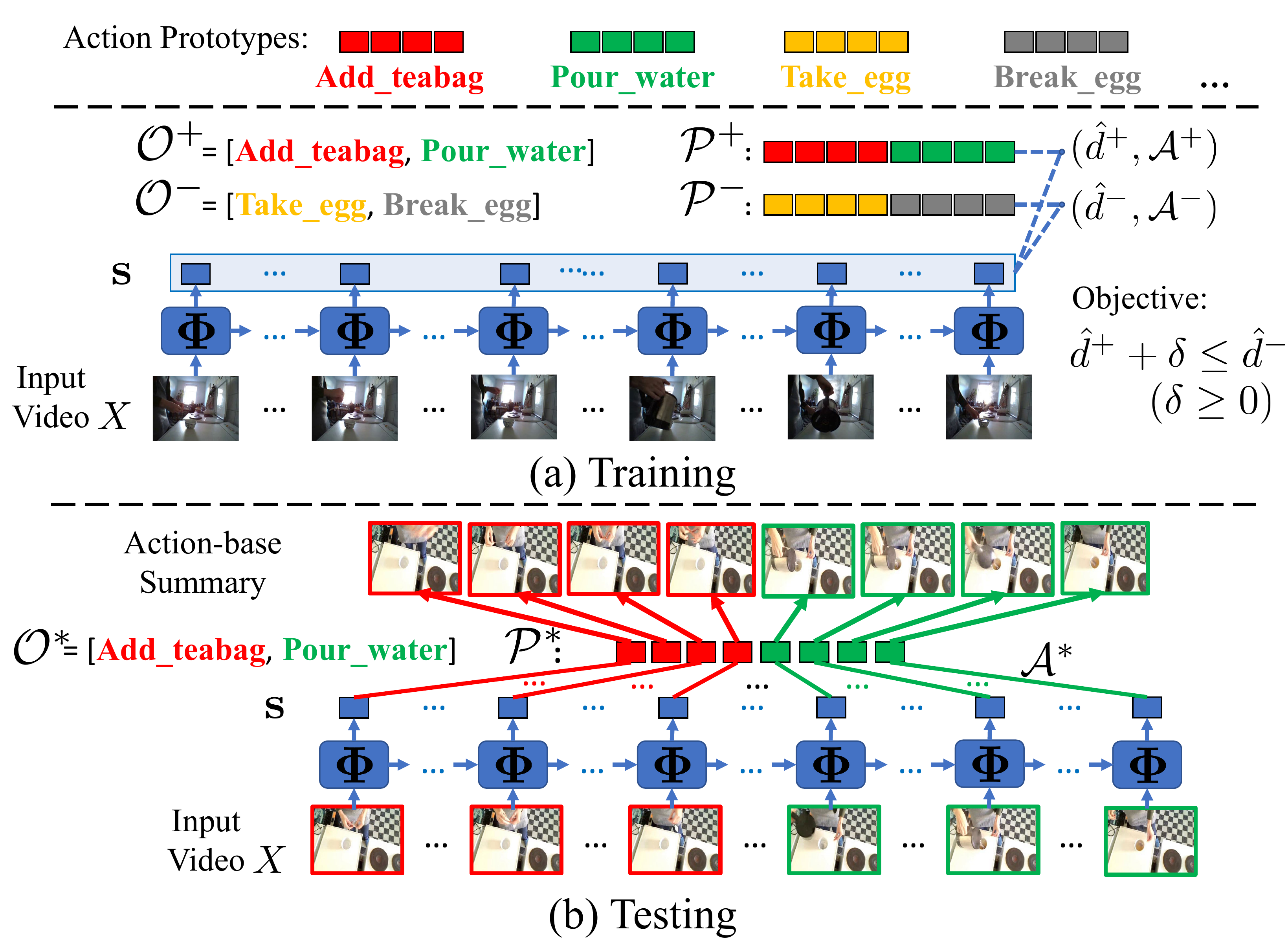}
\end{center}
   \caption{DP-DTW for weakly supervised action segmentation. Each action, indicated by a color, is represented by a prototype sequence with temporal length 4. The frame-wise deep representation $\mathbf{s}$ of input video $X$ is extracted by $\Phi$, e.g., GRU.
   An action transcript $\mathcal{O}$ has its ordering sequence $\mathcal{P}$.
   Evaluated by the DTW discrepancies, a training sample should be closer to its ground truth $\mathcal{O}^+$ than a negative transcript $\mathcal{O}^-$. Hinge loss is used as the discriminative objective.
   The testing transcript $\mathcal{O}^*$ is retrieved or given with its sequence $\mathcal{P}^*$. 
   Based on the DTW alignment $\mathcal{A}^*$ between $\mathbf{s}$ and $\mathcal{P}^*$, an action segmentation, indicated by the colored box on the frame, is obtained. Moreover, the action-based key frames are selected as a video summary.
   }
\label{fig:seg_sum_illus}
\end{figure}

More importantly, DP-DTW can handle temporal recognition of a sequence of multiple classes, as in the weakly supervised action segmentation problem.
Specifically, each action is a class and has its prototype sequence in DP-DTW.
An input video can contain multiple actions performed one after another. Only the action ordering is recorded in the transcript as weak supervision.
By concatenating the action prototypes in order, each transcript has its ordering sequence representation in DP-DTW.
During training, discriminative losses are applied on the DTW discrepancies between the deep representation of the input video and the ordering sequences. As a result, the discriminative action prototypes are learned.
With the retrieved or given transcript of a testing video, the action segmentation is obtained based on the DTW alignment between the input and ordering sequences. Each frame is assigned to the action (prototype) it aligns with.
As a by-product, action-based key frames can be selected by the learned prototypes and used as a summarization of the input video.
The process of DP-DTW mentioned above is illustrated in Figure~\ref{fig:seg_sum_illus}.

The contributions of the proposed method are three-fold.
(1) DP-DTW learns discriminative class-specific prototypes for TSC.
(2) By modeling each action with a temporal sequence as a prototype, the training and inference of DP-DTW for weakly supervised action segmentation are unified under DTW. With the distinctive action dynamics captured by the learned prototypes, action segmentation can then benefit from an optimal temporal alignment.
(3) Action-based video summarization is obtained as a detailed analysis and by-product of the discriminative prototypes learned by DP-DTW.
DP-DTW is evaluated on different temporal recognition tasks. 
On the TSC benchmarks~\cite{dau2019ucr}, DP-DTW outperforms the competitive DTW baselines.
The effectiveness of DP-DTW on weakly supervised action segmentation is demonstrated by state of the art results on two challenging datasets~\cite{kuehne2014language, bojanowski2014weakly}.
Detailed analysis, i.e., action-based summarization, on such videos is enabled by DP-DTW.



\section{Related Work}\label{Sec:RelateWork}

\noindent\textbf{Dynamic Time Warping (DTW).} DTW~\cite{sakoe1978dynamic, berndt1994using} computes the discrepancy value between two sequences based on their optimal alignment from dynamic programming.
Different DTW variants have been proposed. By relaxing the global alignment constraint in DTW, a local optimal matching algorithm~\cite{sakurai2007stream} can be obtained. Shapelet methods~\cite{ye2009time, grabocka2014learning, ma2020adversarial} aim to capture local discriminative temporal dynamics by learning from pre-segmented sub-sequences. With a prediction ensemble~\cite{bagnall2015time, lines2016hive} from DTW and other domains such as frequency, time series classification (TSC) performance can be further boosted.
DTW also has been adopted for different applications, such as heterogeneous sequence alignment~\cite{everingham2006hello, sankar2009subtitle}, time series forecasting~\cite{vincent2019shape} and temporal pattern transform~\cite{lohit2019temporal}.


\noindent\textbf{Learning Prototypes with DTW.} Prototypes capture global temporal patterns over whole input sequences.
With all training samples as class-specific prototypes, the one nearest-neighbour ($1$-NN) classifier with DTW discrepancy as distance can achieve competitive TSC results~\cite{ding2008querying}. However, such models are not efficient.
DBA~\cite{petitjean2014dynamic} learns a few class-specific prototypes by averaging over the sequences of each class and iterative refinement. Its variant, Soft-DTW~\cite{cuturi2017soft}, smooths the dynamic programming procedure~\cite{mensch2018differentiable} and results in a differentiable loss optimized with mini-batch SGD. However, without considering inter-class variance, the prototypes learned by DBA and Soft-DTW are not discriminative.
Beyond TSC tasks, prototypes can also be found in robust data or mid-level feature extractors, i.e.\ a DTW-layer, as proposed in DTWNet~\cite{cai2019dtwnet, iwana2020dtw}.  The learned prototypes in a DTW-layer can be discriminative but latent, i.e., with no explicit correspondence to a specific class.
None of these methods learns class-specific discriminative prototypes as in the proposed DP-DTW.  Comparisons of different prototype learning methods are shown in Table~\ref{tab:DTWs_cmp}.

\begin{table}[t]
\footnotesize
\begin{center}
\begin{tabular}{ccccc}
\hline
\multirow{2}{*}{} & \multicolumn{2}{c}{Prototype} & \multirow{2}{*}{\begin{tabular}[c]{@{}c@{}}mini-batch\\ SGD\end{tabular}} & \multirow{2}{*}{\begin{tabular}[c]{@{}c@{}}Weak-Sup.\\ Act. Seg.\end{tabular}} \\ \cline{2-3}
                  & Cls-Spec.      & Discri.      &                                                                           &                                                                           \\ \hline
DBA~\cite{petitjean2014dynamic}               &       \checkmark         &        $\times$      &              $\times$                                                             &          $\times$                                                                 \\
Soft-DTW~\cite{cuturi2017soft}          &        \checkmark        &       $\times$       &                \checkmark                                                           &        $\times$                                                                   \\
DTWNet~\cite{cai2019dtwnet}            &        $\times$        &      \checkmark        &              \checkmark                                                             &        $\times$                                                                   \\
D$^3$TW$^\dagger$~\cite{chang2019d3tw}              &       $\times$         &      $\times$        &               \checkmark                                                            &                         \checkmark                                                  \\
DP-DTW            &     \checkmark           &      \checkmark        &          \checkmark                                                                 &          \checkmark                                                                 \\ \hline
\end{tabular}
\end{center}
\caption{Comparisons of different DTW models. Temporal classification is the default task. 'Cls-Spec.' and 'Discri.' stand for 'Class-Specific' and 'Discriminative' correspondingly. 'Weak-Sup. Act. Seg.' means weakly supervised action segmentation and $\dagger$ means D$^3$TW is specified for this task only.}
\label{tab:DTWs_cmp}
\end{table}

\noindent\textbf{Weakly Supervised Action Segmentation.} Instead of labeling the action in every single frame, this setting only provides the action ordering of each video for training.
Existing methods for this challenging task follow similar paradigms, encoding frame-wise action predictions with the action ordering to construct different objectives. 
Relevant models include D$^3$TW~\cite{chang2019d3tw}, where Soft-DTW~\cite{cuturi2017soft} is adopted for encoding and loss computation. However, no prototype sequence is learned by D$^3$TW.
An extended connectionist temporal classification model is proposed in \cite{huang2016connectionist}, with frame-to-frame visual similarity as a regularization for frame labels.  Soft boundary assignment  on the initial frame predictions and iterative optimization are done in \cite{ding2018weakly}.  The Viterbi algorithm is another encoding option \cite{richard2018neuralnetwork}. To combine a hidden markov model with deep networks, a loss is proposed by discriminating the energy of all possible frame labelings~\cite{li2019weakly}.
The proposed DP-DTW is different from existing methods in two aspects.  Action prototype sequences are learned by our method and without frame-wise action prediction. As a by-product of the learned prototypes, action-based video summarizations can be obtained.

\noindent\textbf{Video Summarization.} To identify the key frames in a video as a summarization, existing methods assess diversity and/or representativeness and pick distinctive frames.
Unsupervised learning methods~\cite{shemer2019ils, zhou2017deep, zhao2014quasi} use selection criteria from intrinsic temporal structures. 
With explicit annotation of key frames, supervised learning algorithms~\cite{jiang2019comprehensive, fajtl2018summarizing} have been developed.
Under the weakly supervised setting, the category label of each video is provided and used as privileged information to improve summarization \cite{chen2019weakly, cai2018weakly}.
DP-DTW can obtain a type of implicit summarization as a by-product of its alignment process.

\section{Methodology}\label{Sec:Method}

We develop a method for learning discriminative prototypes. We start by providing a recap of the Dynamic Time Warping (DTW) mechanism. Subsequently, the details of our DP-DTW are described.


\subsection{Preliminaries}

\begin{figure}[t]
\begin{center}
  \includegraphics[width=0.9\linewidth]{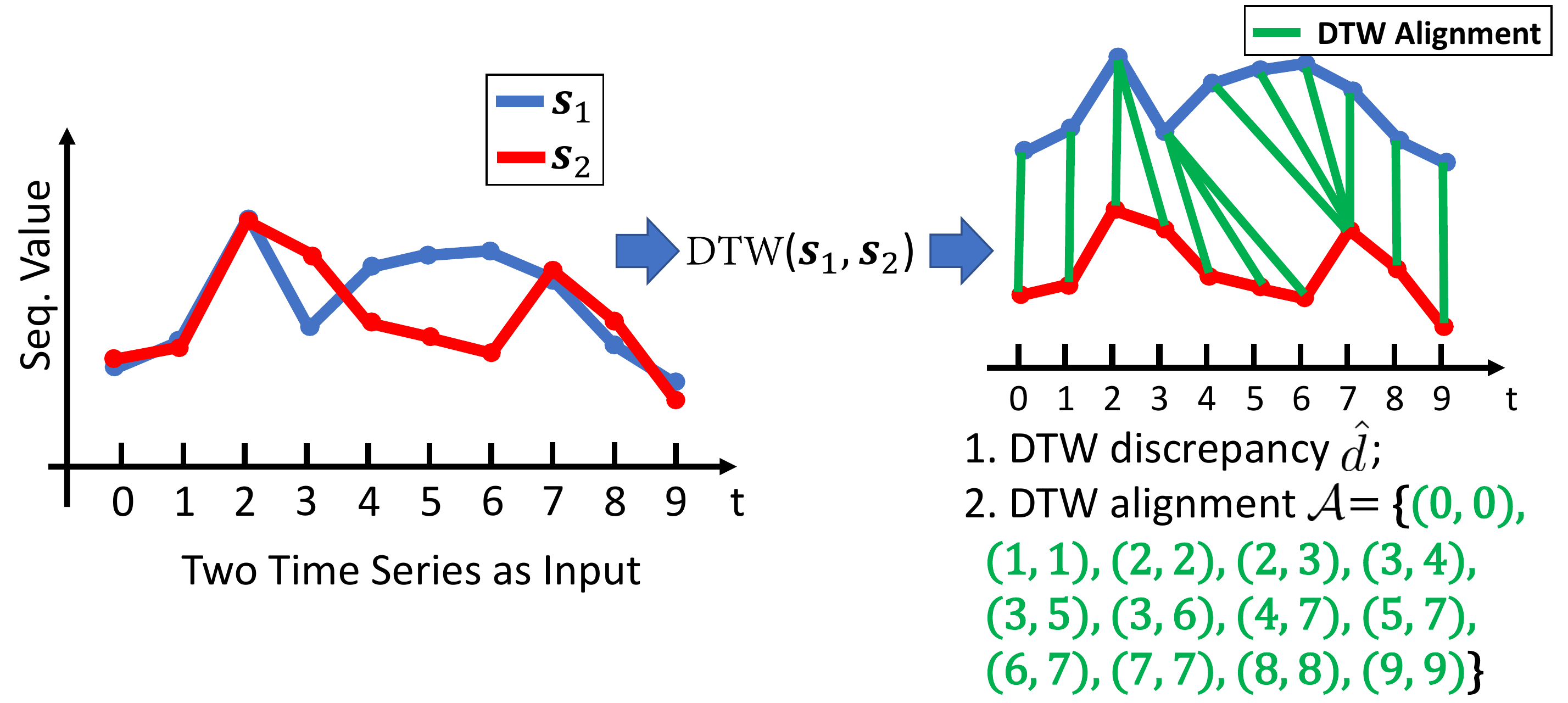}
\end{center}
  \caption{An illustration of DTW inputs and outputs. For example, the ninth alignment $a_9 = (5, 7)$ indicates the aligned moments of $\mathbf{s}_1[5]$ and $\mathbf{s}_2[7]$.}
\label{fig:DTW_toy_illus}
\end{figure}

\noindent\textbf{DTW Mechanisms.} DTW can be treated as a function that takes in two temporal sequences and returns their optimal temporal alignment and the corresponding discrepancy. An input sequence $\mathbf{s} \in \mathbb{R}^{m \times \tau}$ has feature dimension $m$ and temporal length $\tau$. $\mathbf{s}[t] \in \mathbb{R}^{m}$ indicates the feature at time step $t$. Given two sequences $\mathbf{s}_1 \in \mathbb{R}^{m \times \tau_1}$ and $\mathbf{s}_2 \in \mathbb{R}^{m \times \tau_2}$, DTW can be expressed as, 
\begin{equation} \label{eq:DTW_overall}
\mathcal{A}, \hat{d} = \operatorname{DTW}(\mathbf{s}_1, \mathbf{s}_2),
\end{equation}
where $\mathcal{A}$ represents the temporal alignment between ($\mathbf{s}_1$, $\mathbf{s}_2$) and $\hat{d}$ is the discrepancy value. The temporal alignment $\mathcal{A}$ is an optimal solution from dynamic programming and obeys the DTW warping constraints~\cite{berndt1994using}, e.g., monotonicity and continuity.
$\mathcal{A}$ is a list with length $max(\tau_1, \tau_2) \leq |\mathcal{A}| \leq \tau_1 + \tau_2 -1$,
\begin{equation} \label{eq:DTW_align}
\mathcal{A} = \{ a_1, ..., a_{|\mathcal{A}|} \},
\end{equation}
where $a_i = (t_{1,i}, t_{2,i}), i \in \{1, ..., |\mathcal{A}|\}$ indicates the $i$th alignment between $\mathbf{s}_1[t_{1,i}]$ and $\mathbf{s}_2[t_{2,i}]$.
The DTW discrepancy value $\hat{d}$ accumulates along the aligned distances,
\begin{equation} \label{eq:DTW_discrepancy}
\hat{d} = \sum_{i=1}^{|\mathcal{A}|} || \mathbf{s}_1[t_{1,i}] - \mathbf{s}_2[t_{2,i}] ||_2 .
\end{equation}
A toy example of DTW, with $m=1$, $\tau_1=10$, $\tau_2=10$, $|\mathcal{A}|=13$, is illustrated in Figure~\ref{fig:DTW_toy_illus}.

\noindent\textbf{Notation in DP-DTW.} Based on DTW, the proposed DP-DTW aims to learn class-specific discriminative prototypes for temporal recognition tasks.  We assume $\mathcal{K}$ classes are defined in the problem and a set of input sequences with $\mathcal{N}$ samples is denoted as $\mathcal{S} = \{\mathbf{s}_1, \ldots, \mathbf{s}_{\mathcal{N}}\}$. An input sample $\mathbf{s}_n \in \mathbb{R}^{m \times \tau_n}$, $n \in \{1, ..., \mathcal{N}\}$, has feature dimension $m$ and temporal length $\tau_n$ (temporal lengths across inputs can be different).
DP-DTW learns one prototype for each class\footnote{This is for simplicity of presentation. DP-DTW can also support multiple prototypes for each class.}. For a specific class $k\in \{1, ..., \mathcal{K}\}$, its corresponding action prototype is $\mathbf{p}^k \in \mathbb{R}^{m \times \tau_{p}}$, with temporal length fixed at $\tau_{p}$ across different prototypes. The feature of a temporal sequence at time step $t$ is indexed by $\mathbf{s}_n[t]$ or $\mathbf{p}^k[t]$.
Moreover, DP-DTW is optimised with mini-batch SGD. To simplify the derivation, one input sequence $\textbf{s}_n$ is considered.  Generalization to other batch sizes is straightforward.

\subsection{DP-DTW for TSC}
To capture the distinctive temporal dynamics of different classes, DP-DTW focuses on optimizing the inter-class distance between an input sequence and different class-specific prototype sequences.
In the time series classification (TSC) setting, an input sequence $\mathbf{s}_n$ comes from a single class and is labelled with $y_n \in \{1, ..., \mathcal{K}\}$.
For each input $\mathbf{s}_n$, DP-DTW computes its DTW outputs with each learned class-specific prototype $\mathbf{p}^k$, $k\in \{1, ..., \mathcal{K}\}$,
\begin{equation} \label{eq:DTW_TSC}
\mathcal{A}^{k}_{n}, \hat{d}^{k}_{n} = \operatorname{DTW}(\mathbf{p}^k, \mathbf{s}_n).
\end{equation}
In order to use these for classification, a softmax function is applied on the \emph{negative} discrepancy values of all different classes for the logits $\sigma^{k}_{n}$, $k\in \{1, ..., \mathcal{K}\}$.

For learning the prototypes, we include two objectives.
First, a cross entropy loss $\mathcal{L}_{CE} = - \log(\sigma^{y_n}_{n})$ is used, for the purpose of enlarging inter-class distances from prototypes, leading to correct classification.
Moreover, we wish to ensure each prototype represents a class well. Hence, the discrepancy values $\hat{d}^{y_n}_{n}$ form another loss $\mathcal{L}_{D}$.
Therefore, the overall loss function $\mathcal{L}_{TSC}$ consists of two parts,
\begin{equation} \label{eq:DTW_TSC_loss}
\begin{split}
\mathcal{L}_{TSC} & = - \log(\sigma^{y_n}_{n}) + \lambda \cdot \hat{d}^{y_n}_{n} \\
                  & = \mathcal{L}_{CE} + \lambda \cdot \mathcal{L}_{D},
\end{split}
\end{equation}
with a balancing hyper-parameter $\lambda \geq 0$.



\noindent\textbf{Optimization.} 
DP-DTW aims to learn a discriminative prototype for each class via the objective:
\begin{equation} \label{eq:DTW_TSC_obj}
\min_{\{\mathbf{p}^1, ..., \mathbf{p}^{\mathcal{K}}\}} \mathcal{L}_{TSC},
\end{equation}
$\mathcal{L}_{TSC}$ is differentiable and optimized with mini-batch SGD.

\noindent\textbf{Inference.} Once the prototypes $\mathbf{p}^k, k \in \{1, .., \mathcal{K}\}$ are learned against Eq.~\ref{eq:DTW_TSC_obj}, the predicted class label $\tilde{y}$ of a testing input $\mathbf{s}$ is obtained by using the $1$ nearest neighbour (1-NN) classifier with DTW discrepancy,
\begin{equation}
\tilde{y} = \argmin_{k \in \{1, ..., \mathcal{K}\}} \hat{d}^{k}.
\end{equation}
With one discriminative prototype per class learned by DP-DTW, its inference is much more efficient than the baseline 1-NN classifier in which all training samples must be matched to a test sample.

\subsection{DP-DTW for Weakly Supervised Action Segmentation}\label{sec:DPDTW_WeakSupAct}

In weakly supervised action segmentation, only the action ordering, not frame-level labeling, is provided.
By modeling each action as a prototype sequence, the action ordering can also be represented as a sequence in DP-DTW.
Based on the discrepancy and temporal alignment between the video and ordering sequences, the training and inference of DP-DTW are unified under DTW.
Specifically, a discriminative objective is built on the DTW discrepancy for the learning of action prototypes.
The action segmentation can then be inferred and benefit from the optimal alignment by DTW.
Moreover, with the learned action prototypes, action-based video summarization can be obtained by DP-DTW as a by-product.


To handle a video input, a deep neural network (DNN) can be used to extract its feature sequence.
Let $X_n$ denote the raw frames or pre-computed spatial-temporal features of the $n$-th video with $\tau_n$ as its temporal length. A DNN $\Phi(\cdot; \theta)$ is exploited to extract a frame-wise deep representation $\mathbf{s}_n$ from $X_n$,
\begin{equation} \label{eq:DNN_feat_extract}
\mathbf{s}_n = \Phi(X_n; \theta),
\end{equation}
where $\theta$ denotes the DNN model parameters. A deep representation $\mathbf{s}_n[t]$ corresponds to the exact moment of input $X_n[t]$, $t \in \{1, ..., \tau_n\}$.
Moreover, an action ordering can be represented by its corresponding sequence in DP-DTW. Denoting the transcript of the $n$-th sample as $\mathcal{O}_{n} = [o_{n,1}, ..., o_{n,l_n}]$ with $l_n = |\mathcal{O}_{n}|$ actions recorded. Each $o_{n,i}$, $i \in \{1, ..., l_n\}$, specifies the $i$-th action appearing in the video.
Assuming the $i$-th action is $k \in \{1, ..., \mathcal{K}\}$, $o_{n,i} = k$, and it thus corresponds to the action prototype $\mathbf{p}^{k}$.
Therefore, the ordering sequence $\mathcal{P}_n$ is generated by concatenating the action prototypes $\{\mathbf{p}^{1}, ..., \mathbf{p}^{\mathcal{K}}\}$ according to the action ordering recorded in the transcript $\mathcal{O}_{n}$.
This procedure is denoted as $\Pi$,
\begin{equation} \label{eq:prototypes_cat}
\begin{split}
    \mathcal{P}_n & = \Pi(\mathcal{O}_{n}; \{\mathbf{p}^{1}, ..., \mathbf{p}^{\mathcal{K}}\})\\
                  & = \operatorname{TempCat}([\mathbf{p}^{o_{n,1}}, ..., \mathbf{p}^{o_{n,l_n}}]),
\end{split}
\end{equation}
where $\operatorname{TempCat}(\cdot)$ concatenates the list of action prototypes on their temporal dimension and returns the sequence $\mathcal{P}_n \in \mathbb{R}^{m \times \Gamma}$ with temporal length $\Gamma = l_n \cdot \tau_p$.


\noindent\textbf{Training.}
To learn action prototypes with ordering only as weak supervision, both the positive and negative action transcripts are required in building the discriminative learning objective.
For the $n$-th training sample, its ground-truth action ordering is recorded by its positive transcript $\mathcal{O}^{+}_n$ while all other orderings different from the ground-truth can be in its negative transcripts.
However, it is not feasible to consider all possible negative ones, and most of them are not meaningful or seldom appear.
Therefore, a reference ordering set $\mathcal{R}$ is constructed by aggregating all positive transcripts from the training split and keeping the unique orderings.
The feasible negative set of the $n$-th sample is $\mathcal{R} \setminus \mathcal{O}^{+}_n$.
Instead of considering all of them, $Q$ different negative transcripts, $\{\mathcal{O}^{-, 1}_n, ..., \mathcal{O}^{-, Q}_n\} \sim \mathcal{R} \setminus \mathcal{O}^{+}_n$, are randomly selected at each training step for efficiency.
Ordering sequences $\mathcal{P}^{+}_n$ and $\mathcal{P}^{-, q}_n$, $q \in \{1, ..., Q\}$, of the positive and negative transcripts are from Eq.~\ref{eq:prototypes_cat}.
The DTW outputs between $\mathbf{s}_n$ and $\mathcal{P}^{+}_n$ or different $\mathcal{P}^{-, q}_n$s are then computed,
\begin{equation} \label{eq:DTW_pos_neg_dtws}
    \begin{cases}
      \mathcal{A}^{+}_{n}, \hat{d}^{+}_{n} &= \operatorname{DTW}(\mathcal{P}^{+}_n, \mathbf{s}_n);\\
      \mathcal{A}^{-, q}_{n}, \hat{d}^{-, q}_{n} &= \operatorname{DTW}(\mathcal{P}^{-, q}_n, \mathbf{s}_n), q \in \{1, ..., Q\}.
    \end{cases}       
\end{equation}



As a discriminative model, the input sequence $\mathbf{s}_n$ should be closer to its positive ordering prototype than any negative one and formulated as,
\begin{equation} \label{eq:DTW_pos_neg_dists}
   \hat{d}^{+}_{n} < \hat{d}^{-, q}_{n}, \forall q \in \{1, ..., Q\}.
\end{equation}
A hinge loss with margin $\delta \geq 0$ is thus used as the discriminative loss for model training,
\begin{equation} \label{eq:DTW_action_h_loss}
   \mathcal{L}_{h} = \sum_{q=1}^{Q} \max(0, \hat{d}^{+}_{n} - \hat{d}^{-, q}_{n} + \delta).
\end{equation}
Moreover, with $\mathcal{P}^{+}_n$ as a prototype sequence, the $\hat{d}^{+}_{n}$ should be reduced to shrink the representation variance around it, which leads to a distance loss,
\begin{equation} \label{eq:DTW_action_D_loss}
   \mathcal{L}_{D} = \hat{d}^{+}_{n}.
\end{equation}
The overall loss for weakly supervised action segmentation is denoted as $\mathcal{L}_{w\_seg}$,
\begin{equation} \label{eq:DTW_weak_seg_loss}
   \mathcal{L}_{w\_seg} = \mathcal{L}_{h} + \lambda \cdot \mathcal{L}_{D},
\end{equation}
with a balancing hyper-parameter $\lambda \geq 0$.

\noindent\textbf{Optimization.} The loss $\mathcal{L}_{w\_seg}$ is built on deep representation $\mathbf{s}_n$ and action prototypes $\{\mathbf{p}^1, ..., \mathbf{p}^{\mathcal{K}}\}$. $\mathbf{s}_n$ is extracted by deep model $\Phi(\cdot; \theta)$ from raw video input $X_n$. Therefore, the learning objective is defined as,
\begin{equation} \label{eq:DTW_weak_sup_act_seg_obj}
\min_{\{\mathbf{p}^1, ..., \mathbf{p}^{\mathcal{K}}\}; \theta} \mathcal{L}_{w\_seg},
\end{equation}
where the action prototypes and DNN parameters are jointly optimized with mini-batch SGD.

\noindent\textbf{Inference.} In the \emph{segmentation} setting, only the testing raw input $X_n$ is available. Its frame-wise deep representation $\mathbf{s}_n$ is extracted by the learned DNN $\Phi$ as in Eq.~\ref{eq:DNN_feat_extract}. Due to the exact temporal correspondence between $\mathbf{s}_n$ and $X_n$, the analysis, e.g., alignments, on $\mathbf{s}_n$ can be easily tracked back to the raw frames in $X_n$.

The best matching transcript $\mathcal{O}^*_n$ is retrieved from the reference set $\mathcal{R}$ by,
\begin{equation} \label{eq:DTW_seg_get_order}
\mathcal{O}^*_n = \argmin_{\mathcal{O} \in \mathcal{R}} \operatorname{DTW}(\Pi(\mathcal{O}; \{\mathbf{p}^{1}, ..., \mathbf{p}^{\mathcal{K}}\}), \mathbf{s}_n),
\end{equation}
where $\Pi$ is the concatenation procedure defined in Eq.~\ref{eq:prototypes_cat} and $\argmin$ compares DTW discrepancies only. $\mathcal{P}^*_n$ is the ordering sequence of $\mathcal{O}^*_n$.
The action assignment on the input sequence $\mathbf{s}_n$ can then be obtained based on the optimal alignment $\mathcal{A}^{*}_{n}$ by DTW,
\begin{equation}
\mathcal{A}^{*}_{n}, \hat{d}^{*}_{n} = \operatorname{DTW}(\mathcal{P}^{*}_n, \mathbf{s}_n).
\end{equation}
Specifically, $a_i = (t_{1,i}, t_{2,i})$ is the $i$th alignment of $\mathcal{A}^{*}_{n}$ (as in Eq.~\ref{eq:DTW_align}) and it indicates the aligned pair of $\mathcal{P}^{*}_n[t_{1,i}]$ and $\mathbf{s}_n[t_{2,i}]$. The action of $\mathbf{s}_n$ at time step $t_{2,i}$ is consistent with the action of $\mathcal{P}^{*}_n$ at $t_{1,i}$, which can be easily determined with $\mathcal{O}^*_n$ and the action prototype length $\tau_p$. Moreover, a frame in $\mathbf{s}_n$ can align with multiple continuous steps in $\mathcal{P}^{*}_n$ and its nearest neighbour is chosen for action label assignment.
In the \emph{alignment} setting, the ground-truth transcript $\mathcal{O}^+_n$ is given. It can be handled with a similar procedure to that described above by replacing $\mathcal{O}^*_n$ with $\mathcal{O}^+_n$ accordingly.



\subsubsection{Action-based Video Summarization}\label{sec:DPDTW_WeakSupAct_sum}
With the learned discriminative prototypes of different actions $\{\mathbf{p}^{1}, ..., \mathbf{p}^{\mathcal{K}}\}$, a summarization of the input video can be obtained by aggregating the key moments of each action according to the transcript.
Specifically, the ground-truth action transcript $\mathcal{O}^+_n$ is provided along with the video input $X_n$. $\mathcal{P}^+_n$ is the ordering sequence of $\mathcal{O}^+_n$. $\mathcal{P}^+_n \in \mathbb{R}^{m \times \Gamma}$ is with temporal length $\Gamma = l_n \cdot \tau_p$, where $l_n$ is the number of actions appearing in the transcript $\mathcal{O}^+_n$ and $\tau_p$ is the temporal length of each action prototype.
Based on the DTW alignment $\mathbf{A}^+_n$ between $\mathbf{s}_n$ and $\mathcal{P}^+_n$, each $\mathcal{P}^+_n[t]$, $t \in \{1, ..., \Gamma\}$, has a nearest neighbour from its aligned $\mathbf{s}_n$ and is denoted as $\mathbf{s}_n[t^{\prime}]$. $t^{\prime}$ is treated as one of the key or representative moments of the input sequence and there are $\Gamma$ key moments in total. In the ideal case, every $\tau_p$ key moments under the same action corresponds to the given transcript $\mathcal{O}^+_n$. Therefore, the selected $\Gamma$ key frames from $X_n$ are the action-based summarization of the video.

\section{Experiments}\label{Sec:Exp}

The main characteristic of DP-DTW is computing class-specific discriminative prototypes for temporal recognition tasks. In this section, the effectiveness of the proposed model is verified on time series classification (TSC) and weakly supervised action segmentation. Moreover, with the prototypes learned by DP-DTW, we show that detailed analysis, i.e., summarizing the input videos with action-based key frames, can be achieved.

\subsection{TSC}

\noindent\textbf{Dataset.} \textbf{UCR}~\cite{dau2019ucr} is a benchmark collection of 128 univariate time series datasets with different application backgrounds such as electronics and biology. The sequence temporal length is the same within each dataset. Each dataset has multiple classes and each sequence belongs to one class.

\noindent\textbf{Implementation Details.} In each UCR dataset, the number of prototypes $\mathcal{K}$ in DP-DTW is set as the number of classes in the dataset, i.e.\ one prototype per class. The temporal length $\tau_p$ of each prototype is fixed as the input sequence length. The feature dimension $m$ of both input and prototype sequences is 1 in the UCR datasets.
DP-DTW is directly applied on the raw input sequences of each UCR dataset to learn the discriminative prototypes of classes. The prototypes are initialized with the medoids of different classes.
20$\%$ of the training data in each dataset forms a mini-batch and 60 epochs are used for learning. We optimise with Adam~\cite{kingma2014adam} and cross-validate the hyper-parameters such as learning rate.

\noindent\textbf{Competitors.} 
The proposed DP-DTW is compared with the four baselines:
(1) \textit{1-NN + ED} is the 1 nearest neighbour (1-NN) classifier with Euclidean distance (ED) for TSC.
(2) \textit{1-NN + DTW} is 1-NN with DTW discrepancy for TSC.
(3) \textit{1-NN + DTW(W)} is 1-NN with DTW discrepancy (by tuning its window size constraint to optimal) for TSC.
(4) \textit{DBA}~\cite{petitjean2014dynamic} averages the time series within each class as prototype.
The first three methods are the common baselines of UCR datasets with their results listed on the project page\footnote{\url{https://www.cs.ucr.edu/~eamonn/time_series_data_2018/}, we also reproduce them as a sanity check.}. Moreover, the first three 1-NN based methods are strong baselines with all training samples as reference. On the contrary, the prototype learning methods, e.g., DP-DTW and DBA, are with one prototype learned for each class.

\begin{figure}[t]
\begin{center}
  \includegraphics[width=0.95\linewidth]{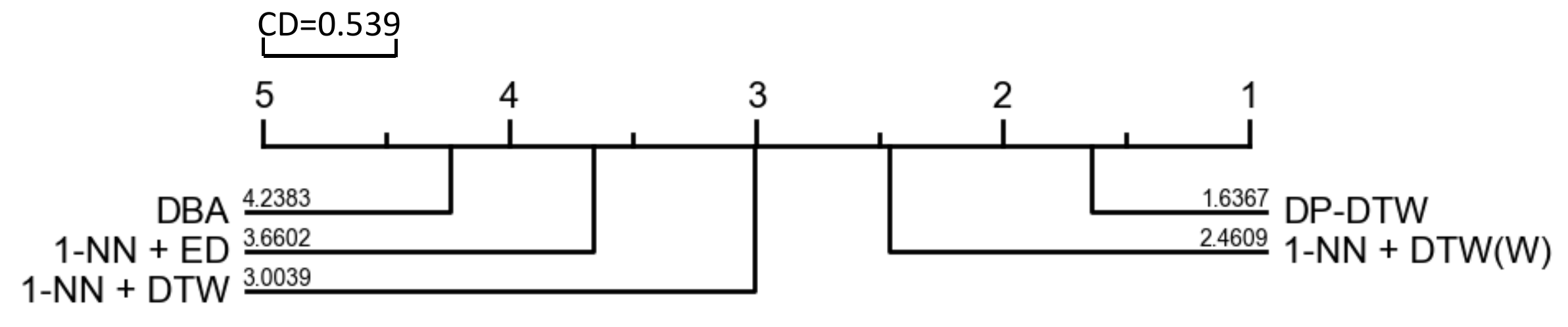}
\end{center}
  \caption{Critical difference diagram on the UCR 128 datasets with five TSC algorithms compared. The critical difference (CD) is $0.539$ with significance level at $0.05$.}
\label{fig:TSC_CDD_rank}
\end{figure}

\begin{figure}[t]
\begin{center}
  \includegraphics[width=0.9\linewidth]{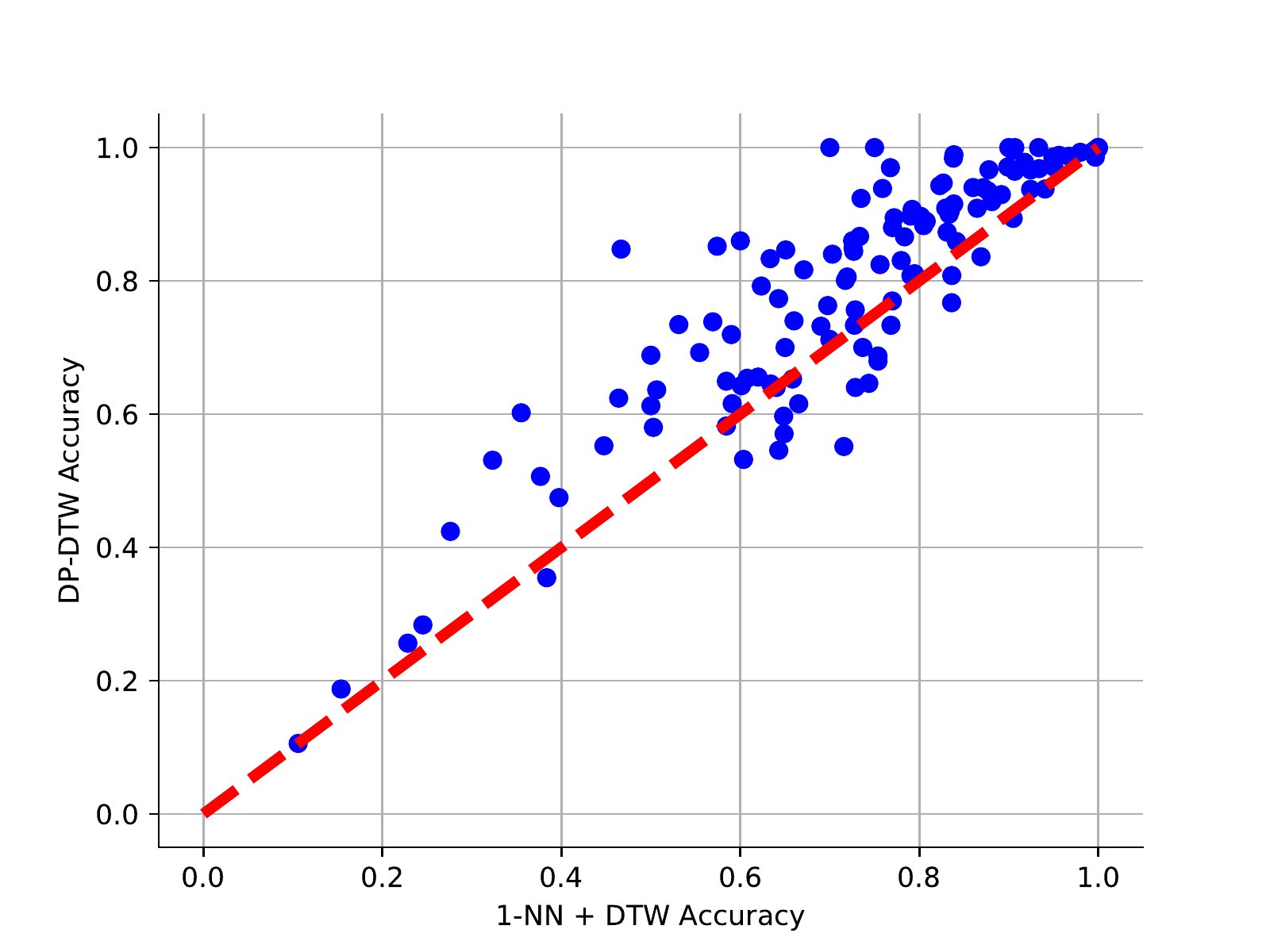}
\end{center}
  \caption{Each point corresponds to two TSC test accuracies on a UCR dataset by DP-DTW (y-axis) and 1-NN + DTW (x-axis). The points above or on the diagonal red dashed line mean DP-DTW achieves no worse results than 1-NN + DTW on such datasets. This is the case for 82.8$\%$ of the UCR datasets.}
\label{fig:TSC_DTWvs1NN}
\end{figure}

\noindent\textbf{Results.} 
To collectively compare classification performance of the five TSC methods over the $128$ datasets, critical difference diagram~\cite{demvsar2006statistical} can be exploited, as shown in Figure~\ref{fig:TSC_CDD_rank}.
The averaged (over 128 datasets) ranks of different classifiers are compared. The better method is with the lower rank.
DP-DTW is thus the best model among them with the lowest averaged rank ($\sim$1.6) achieved. Different 1-NN classifiers come after with higher ranks.
Moreover, the differences among these methods are statistically significant as the gaps of their averaged ranks are all greater than the critical difference (CD) $0.539$.
For more targeted analysis, different methods can be compared in a pair. A model that achieves superior results to its competitor on the majority of datasets is the better one.
The pairwise comparisons between our DP-DTW and the four competitors are as follows. DP-DTW achieves no worse results than 1-NN + ED on $86.7\%$ $(111/128)$ of all datasets. Comparing with the DTW based 1-NN classifiers, DP-DTW still has no worse classification performance on $82.8\%$ $(106/128)$ and $74.2\%$ $(95/128)$ of all datasets to 1-NN + DTW and 1-NN + DTW(W) respectively.
We also compare DP-DTW with another competitive baseline, shapeDTW~\cite{zhao2018shapedtw}, and the former one is $70.3\%$ better than the latter one\footnote{10 out of 128 UCR datasets took shapeDTW too long to finish and thus excluded from comparison.}.
A scatter plot is also used to show the details of comparison between DP-DTW and 1-NN + DTW, as in Figure~\ref{fig:TSC_DTWvs1NN}.
DP-DTW clearly outperforms the averaging based DBA~\cite{petitjean2014dynamic} with no worse results on all 128 datasets.
Both the critical difference diagram of ranking and pairwise comparisons demonstrate the importance of learning discriminative prototypes for different classes.

\begin{table}[t]
\footnotesize
\begin{center}
\begin{tabular}{c|cccccc}
\hline
       & \multicolumn{3}{c}{Breakfast} & \multicolumn{3}{c}{Hollywood} \\
       & F-acc.     & IoU     & IoD     & F-acc.     & IoU    & IoD     \\ \hline
HMM+RNN~\cite{richard2017weakly}   &     33.3       &    -     &    -     &     -   &   11.9     &    -     \\
TCFPN~\cite{ding2018weakly}   &      38.4      &     24.2    &    40.6     &     28.7   &   12.6     &     18.3    \\
NN-Viterbi~\cite{richard2018neuralnetwork}   &     43.0       &    -     &     -    &    -        &   -     &    -     \\
D$^3$TW~\cite{chang2019d3tw}   &     45.7       &    -     &    -     &      33.6  &    -    &     -    \\
CDFL~\cite{li2019weakly}   &      50.2      &     33.7    &    \textbf{45.4}     &      45.0      &    19.5    &    25.8     \\ \hline
DP-DTW &      \textbf{50.8}      &     \textbf{35.6}    &     45.1    &      \textbf{55.6}      &    \textbf{33.2}    &     \textbf{43.3}    \\ \hline
\end{tabular}
\end{center}
\caption{Comparisons among DP-DTW and competitors under the segmentation setting.}
\label{tab:weak_sup_seg_results}
\end{table}

\begin{figure}[t]
\begin{center}
   \includegraphics[width=0.85\linewidth]{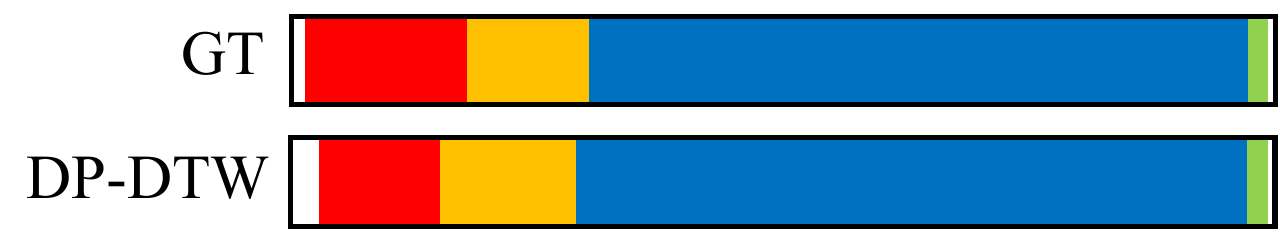}
\end{center}
   \caption{The illustrated test video is 'P39\_cam02\_P39\_friedegg' from Breakfast dataset. The retrieved transcript is the same as ground-truth (GT). Different actions are represented by different colors, i.e., \textcolor{red}{butter\_pan}, \textcolor[rgb]{1.0,0.753,0.}{crack\_egg}, \textcolor{blue}{fry\_egg}, \textcolor{green}{put\_egg2plate}, and white indicates background. Best viewed in color.}
\label{fig:DPDTW_Seg_GT_CMP}
\end{figure}

\subsection{Weakly Supervised Action Segmentation}

\noindent\textbf{Datasets.} 
\textbf{Breakfast}~\cite{kuehne2014language} contains 1,712 videos of 48 actions related to the preparation of breakfast. Such videos are recorded from multiple views under real-life scenarios, e.g., from 18 different home kitchens. On average, $\sim$7 action instances are performed one after another in each video.
We follow the data splits in \cite{kuehne2014language} and the averaged results are reported.
\textbf{Hollywood Extended}~\cite{bojanowski2014weakly} comprises 16 action classes and 937 videos clipped from Hollywood movies. There are 2 to 11 actions in each video and 2.5 actions on average. For a fair comparison, the pre-computed frame-level features and the data split criteria in \cite{richard2017weakly, richard2018neuralnetwork} are used.
In both datasets, the action ordering of each video is recorded in a transcript as weak supervision.

\noindent\textbf{Settings and Metrics.} \emph{Segmentation} and \emph{alignment} are two sub-tasks in weakly supervised action segmentation.
In the segmentation task, the ground-truth action transcript is not available during evaluation. The best matching action ordering is first retrieved from the reference set $\mathcal{R}$. The label assignment procedure is then applied.
In the alignment task, the ground-truth transcript is provided. Action labels are assigned following the ground-truth action order.
Three standard metrics are adopted to evaluate both settings.
The first is the frame accuracy (\textbf{F-acc.}), the percentage of frames that are correctly labeled. The other two are the intersection over union (\textbf{IoU}) and the intersection over detection (\textbf{IoD}). Given a ground-truth action assignment $I^*$ and the predicted assignment $I$, $\operatorname{IoU} = |I \cap I^*| / |I \cup I^*|$ and $\operatorname{IoD} = |I \cap I^*| / |I|$.

\begin{table}[t]
\footnotesize
\begin{center}
\begin{tabular}{c|cccccc}
\hline
       & \multicolumn{3}{c}{Breakfast} & \multicolumn{3}{c}{Hollywood} \\
       & F-acc.     & IoU     & IoD     & F-acc.     & IoU    & IoD     \\ \hline
HMM+RNN~\cite{richard2017weakly}   &      -      &    -     &    47.3     &      -  &    -    &    46.3     \\
TCFPN~\cite{ding2018weakly}   &     53.5       &    35.3     &    52.3     &     57.4       &   22.3     &    39.6     \\
NN-Viterbi~\cite{richard2018neuralnetwork}   &      -      &    -     &    -     &    -        &    -    &    48.7     \\
D$^3$TW~\cite{chang2019d3tw}   &      57.0      &    -     &    56.3     &    59.4  &    -    &     50.9    \\
CDFL~\cite{li2019weakly}   &      63.0      &     45.8    &    63.9     &       64.3     &    40.5    &     52.9    \\ \hline
DP-DTW &     \textbf{67.7}       &    \textbf{50.8}     &    \textbf{66.5}     &      \textbf{66.4}      &    \textbf{46.8}    &    \textbf{61.7}     \\ \hline
\end{tabular}
\end{center}
\caption{Comparisons among DP-DTW and competitors under the alignment setting.}
\label{tab:weak_sup_align_results}
\end{table}

\begin{figure}[t]
\begin{center}
   \includegraphics[width=0.85\linewidth]{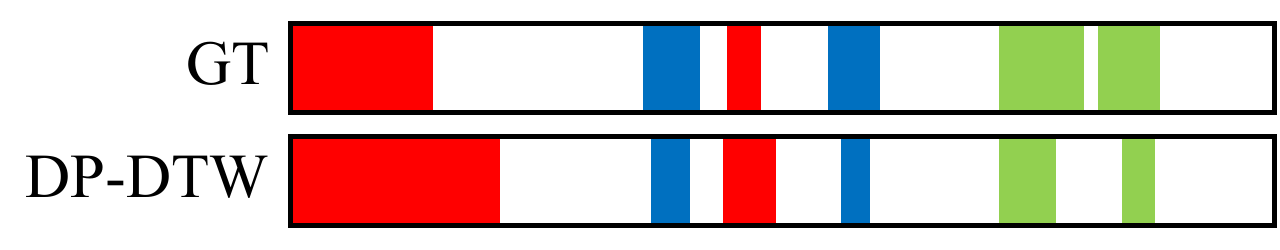}
\end{center}
   \caption{Illustration of video action alignment. The test video consists of multiple actions, i.e., \textcolor{red}{Run}, \textcolor{blue}{SitDown}, \textcolor{green}{StandUp} and white as background. The ground truth action transcript is given. Best viewed in color.}
\label{fig:DPDTW_Align_GT_CMP}
\end{figure}

\begin{figure*}[t]
\begin{center}
   \includegraphics[width=0.8\linewidth]{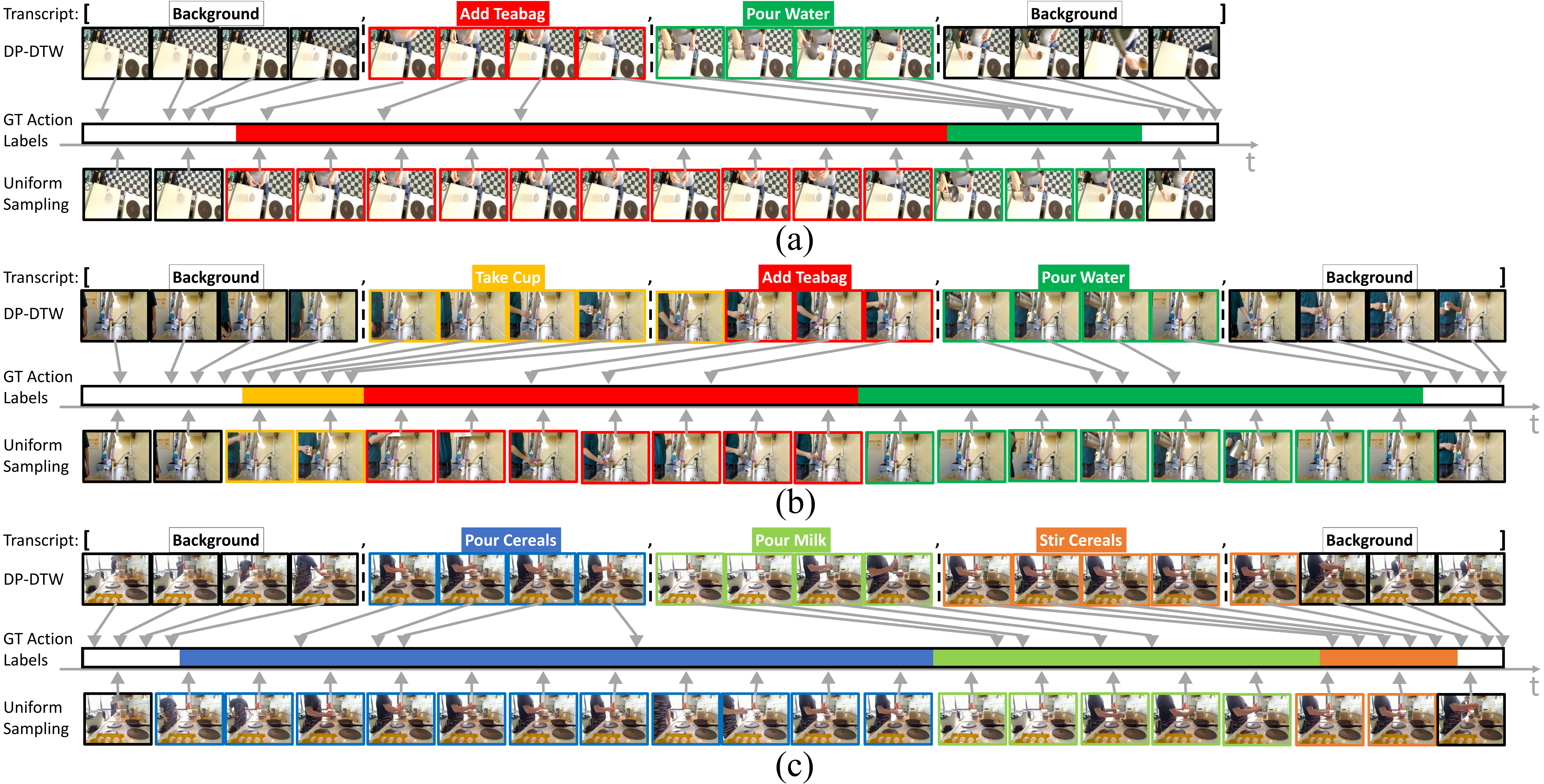}
\end{center}
   \caption{Summarizations of three videos by DP-DTW and uniform sampling. Different actions are indicated by different colors. The length of a color bar reflects the relative duration of the corresponding action in a video.
   Selected moments of the frames are also indicated. The action performed in a selected frame is the same as its selected moment and indicated by its frame box color.
   In DP-DTW summarizations, dashed lines split the summary frames according to the ideal action-based case where the actions of selected frames within each split should be consistent with the action in transcript.}
\label{fig:action_video_sums}
\end{figure*}

\noindent\textbf{Implementation Details.} The deep model $\Phi$ is a single-layer GRU with 256 hidden units. Therefore, the feature dimension $m = 256$.
$\mathcal{K} = 17$ (16 actions plus 1 background class) for the Hollywood dataset and $\mathcal{K} = 48$ (with background class included) for Breakfast. The temporal length $\tau_p$ of prototype is $6$ and $8$ for Hollywood and Breakfast respectively.
The number of randomly selected negative transcripts $Q$ (in Eq.~\ref{eq:DTW_pos_neg_dtws}) is $50$ during training.
The hinge loss margin $\delta = 1$ in Eq.~\ref{eq:DTW_action_h_loss}.
To minimize objective $\mathcal{L}_{w\_seg}$ in Eq.~\ref{eq:DTW_weak_seg_loss} with balancing $\lambda = 0.1$,
an Adam optimizer is used with mini-batch size $64$ and initial learning rate at $0.001$ for $10,000$ training steps.

\noindent\textbf{Competitors.} The state of the art methods for weakly supervised action segmentation are compared to: (1) A new loss is proposed by \textit{CDFL}~\cite{li2019weakly} to discriminate the energy of all valid and invalid action assignment paths.
(2) In \textit{D$^3$TW}~\cite{chang2019d3tw}, the dynamic programming of DTW is exploited as an optimal encoding and a discriminative DTW loss is proposed.
(3) \textit{NN-Viterbi}~\cite{richard2018neuralnetwork} proposes a Viterbi-based loss that enables online learning.
(4) \textit{TCFPN}~\cite{ding2018weakly} is based on the frame-wise label prediction. Network updates are then performed iteratively for better efficiency.
(5) In \textit{HMM+RNN}~\cite{richard2017weakly}, an HMM ensures the assignments obey the action order while an RNN makes the frame-wise predictions.
These methods are mainly based on encoding frame predictions with an action ordering.

\noindent\textbf{Segmentation Results.} The proposed DP-DTW is compared with different competitors under the segmentation setting, as shown in Table~\ref{tab:weak_sup_seg_results}.
DP-DTW achieves one of the best results on the Breakfast dataset. On the Hollywood Extented, it clearly outperforms the state of the art, CDFL~\cite{li2019weakly}, with $10.6\%$, $13.7\%$ and $17.5\%$ improvements on frame accuracy, IoU and IoD respectively.
Comparing with the DTW based method, D$^3$TW~\cite{chang2019d3tw}, DP-DTW also achieves superior performance to it with clear margins, $5.1\%$ better frame accuracy on Breakfast and $22.0\%$ better on Hollywood. These results demonstrate the effectiveness and necessity of explicitly learning discriminative prototypes of different actions for the weakly supervised action segmentation problems.
The qualitative comparison between action segmentation by DP-DTW and ground-truth labels is illustrated in Figure~\ref{fig:DPDTW_Seg_GT_CMP}. The majority of sequential actions with varied temporal lengths can be localized by DP-DTW. However, it is challenging to accurately determine the true start and end moments across actions.

\noindent\textbf{Alignment Results.} Comparing with the segmentation task, the alignment one is less challenging due to the ground-truth action transcripts being provided during testing. As a result, a model's performance under the alignment setting is generally much better than the segmentation one, as shown in Table~\ref{tab:weak_sup_align_results}. DP-DTW achieves the best results on both datasets. Comparing with D$^3$TW~\cite{chang2019d3tw}, DP-DTW still consistently achieves better results with clear ($\geq 7.0\%$) margin over all criteria.
The action alignment of a video ($\textit{0652}$) from Hollywood Extended dataset is illustrated in Figure~\ref{fig:DPDTW_Align_GT_CMP}. Good action alignments can be achieved by DP-DTW even with frequent action transitions.

\subsection{Action-Based Video Summarization}


With the action prototypes learned and the transcripts provided in weakly supervised action segmentation, the action-based key frames can be selected by DP-DTW as a by-product and used to summarize the input video, as detailed in 
Sec.~\ref{sec:DPDTW_WeakSupAct_sum}.
For illustration purposes, a DP-DTW model with prototype temporal length $\tau_p = 4$ is trained on Breakfast. Four key frames are thus selected by each prototype sequence as the summary of an action. In the ideal case, such frames should all belong to the corresponding action and an action-based video summarization can be obtained according to the action transcript.
As shown in Figure~\ref{fig:action_video_sums}, the key frames selected by a prototype are often distinctive moments across the time span of the corresponding action. The DP-DTW summarization is action-based and thus robust to action duration variation in the video.
On the contrary, summarizing a video by uniform sampling can only reflect the duration of different actions.
By comparing the action labels of the selected frames with the ideal action-based summarization case (inferred from the action transcript and $\tau_p$), the matching rate (accuracy) can be obtained. The summarization by DP-DTW achieves $62.5\%$ accuracy while uniform sampling achieves $40.8\%$.

\section{Conclusion}\label{Sec:Conclude}
We proposed Discriminative Prototype DTW (DP-DTW), the first model to explicitly learn class-specific discriminative prototypes for temporal recognition.
Different from existing methods, DP-DTW focuses on enlarging the inter-class difference among prototypes via a discriminative objective.  We develop an algorithm for weakly supervised segmentation based on DP-DTW prototype sequences, with end-to-end learning.  DP-DTW outperforms competitive baselines on TSC benchmarks. On two challenging weakly supervised action segmentation datasets, DP-DTW achieves state of the art results.  Action-based video summarization is also enabled by DP-DTW via alignment to the input video.

{\small
\bibliographystyle{ieee_fullname}
\bibliography{egbib}
}

\end{document}